\documentclass{article}

\usepackage{iclr2027_conference,times}

\usepackage{amsmath,amsfonts,bm}

\def\eqref#1{equation~\ref{#1}}
\def\1{\bm{1}}

\DeclareMathAlphabet{\mathsfit}{\encodingdefault}{\sfdefault}{m}{sl}
\SetMathAlphabet{\mathsfit}{bold}{\encodingdefault}{\sfdefault}{bx}{n}

\usepackage{hyperref}
\usepackage{url}
\usepackage{graphicx}
\usepackage{booktabs}
\usepackage{wrapfig}
\usepackage{multirow}
\usepackage[table]{xcolor}
\usepackage{algorithm}
\usepackage{algorithmic}

\definecolor{firstgreen}{HTML}{2F7D5A}
\definecolor{secondgray}{HTML}{F2F2F2}
\definecolor{shopmindbg}{HTML}{EDF3F0}

\newcommand{\first}[1]{%
  \textcolor{firstgreen}{\textbf{#1}}\rlap{\,$^{\star}$}%
}
\newcommand{\second}[1]{%
  \underline{#1}\rlap{$^{\dagger}$}%
}
\newcommand{\third}[1]{%
  #1\rlap{$^{\ddagger}$}%
}

\iclrfinalcopy

\title{ShopMind: A Mental World Model for Personalized and Realistic E-commerce User Simulation}

\author{
  Peixuan Hou\textsuperscript{1},
  Bin Chen\textsuperscript{1},
  Li He\textsuperscript{2},
  Jian Xu\textsuperscript{2},
  Xiuli Ma\textsuperscript{1},
  Guojie Song\textsuperscript{1}\thanks{Corresponding author.}\\
  \textsuperscript{1}State Key Laboratory of General Artificial Intelligence,\\
  School of Intelligence Science and Technology, Peking University\\
  \textsuperscript{2}Alibaba Group\\
}

\iclrfinalcopy

\begin{document}

\maketitle

\lhead{Preprint}

\begin{abstract}
Large language models are widely used in e-commerce user simulation, providing a foundation for understanding users and predicting human decisions. However, existing simulators still struggle to achieve both personalization and behavioral realism. Most methods predict the next action directly from interaction histories and the current context, without accounting for the evolving mental world that underlies user behavior. Theory of Mind suggests that behavior is driven by mental states, including beliefs, uncertainties, and intentions. Users may interpret the same information in distinct ways, leading to divergent behaviors, while an individual’s mental state evolves through interaction and shapes subsequent actions.
Ignoring this process prevents simulators from faithfully capturing individual differences and realistic behavior patterns, and undermines its effectiveness in downstream applications. To address this, we propose \textbf{ShopMind}, a \textbf{Mental World Model} for e-commerce user simulation that captures how consumers interpret shopping information and how their mental states evolve, better reflecting individual users and real-world behavior patterns. We further use behavior–mental state pairs synthesized by ShopMind to train a \textbf{Mental State Reconstruction (MSR) model}, which infers consumers’ latent mental states from observable behavior. Experiments show that ShopMind improves personalization by 18.0\% and behavioral realism by 19.6\% over the strongest alternatives. Moreover, MSR improves performance on consumption value measurement and CTR prediction, showing the utility of ShopMind for understanding and predicting real user behavior.

%, showing the utility of ShopMind for understanding and predicting real user behavior

\end{abstract}

\section{Introduction}
%Large language models (LLMs) have shown strong potential for simulating human online shopping behavior, making it possible to construct credible virtual consumers. LLM-based e-commerce user simulation can provide low-cost evaluation environments for recommender systems and has broader applications in cognitive science and psychology. It offers an important foundation for understanding human decision-making and improving behavioral prediction.

Large language models (LLMs) have shown strong potential for simulating human online shopping behavior~\citep{aher2023using,argyle2023out,park2023generative,horton2024economic}. LLM-based e-commerce user simulation can provide low-cost evaluation environments for recommender systems~\citep{zhang2025llmuser,huang2025recommender} and has broader applications in cognitive science and psychology~\citep{binz2023using,dillion2023can,lin2026psychological}. It offers an important foundation for understanding human decision-making and improving behavioral prediction.

\begin{figure*}[t!]
    \centering    
    \includegraphics[width=\linewidth, keepaspectratio]{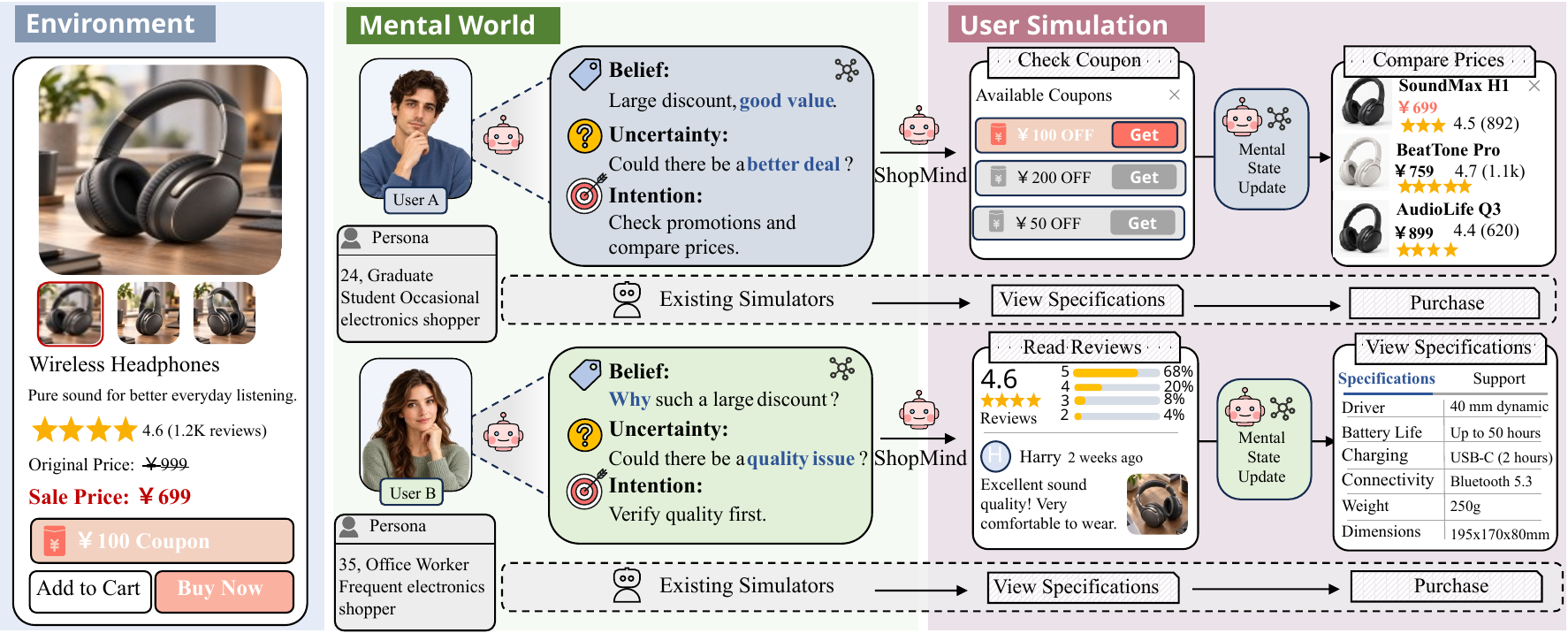}
    \vspace{-1.5em}
    \caption{Illustration of the proposed \textbf{ShopMind}. Users may form different mental states from the same shopping information, leading to different behaviors. By modeling this evolving mental world, ShopMind enables more personalized and realistic user simulation than other approaches.
    }
    \label{fig:teaser}
    \vspace{-1em}
\end{figure*}

Recent studies have increasingly explored LLM-based e-commerce user simulation. 
Given a user profile, interaction history, and the current shopping context, an e-commerce user simulator predicts the user's next action. 
Existing approaches mainly improve simulation quality along two directions: some leverage real shopping trajectories for supervised fine-tuning or reinforcement learning to better align simulated behaviors with real users~\citep{lu2026multiturn, zhang2026shopr1}, while others condition user simulation on personas, preferences, and interaction histories to capture individual behavioral differences~\citep{mansour2025paars, wang2025customerr1, wang2026opera, ling2026shopperbench}. 
However, existing simulators still struggle to achieve both personalized and realistic behavior. 
A key limitation is that they largely predict the next action from observable signals, without explicitly modeling the evolving mental world of consumers throughout the shopping process. 
User behavior cannot be fully explained by observable context alone, as decisions are also shaped by how individuals perceive and interpret the information they encounter~\citep{bettman1998constructive}. 
Theory of Mind provides a natural perspective for this view: Behavior is driven by mental states, including beliefs, intentions, and uncertainties~\citep{premack1978chimpanzee}. 
For example, when exposed to the same discounted product, a price-sensitive user may perceive it as a purchase opportunity, whereas a quality-oriented user may interpret the discount as a signal of lower quality and make a different decision. 
In addition, the same user's mental state may evolve through browsing, further shaping subsequent behavior. Ignoring the mental world limits both the personalization and realism of simulated behavior and undermines the simulator's effectiveness for user understanding and behavioral prediction.
%Ignoring these mental states and their evolution limits a simulator's ability to capture individualized decision processes, undermining both the personalization and realism of simulated behavior and its effectiveness for user understanding and behavioral prediction.

To address this limitation, we propose \textbf{ShopMind}, a \textbf{mental world model} for e-commerce user simulation. As shown in Figure~\ref{fig:teaser}, ShopMind continuously models and updates consumers' beliefs, uncertainties, and intentions to capture how they interpret shopping information and make decisions, enabling more realistic and personalized simulation. It consists of three core modules: \textbf{Mental State Modeling and Transition} tracks the consumer's evolving mental state; \textbf{Mental-World Rollout} explores candidate actions by simulating their subsequent mental-state and behavioral trajectories; and \textbf{Human-Aligned Behavior Selection} selects the next action by balancing individual consistency and behavioral realism, with realism assessed by a Behavior Critic trained on real shopping trajectories. The selected action drives the next interaction, which in turn updates both the shopping environment and the consumer's mental state, yielding shopping trajectories that are both behaviorally plausible and tailored to individual users. Throughout this process, ShopMind continuously aligns evolving mental states with generated behaviors, naturally producing behavior--mental-state pairs that are largely unavailable in real-world interaction logs. We use these synthesized pairs to train a \textbf{Mental State Reconstruction (MSR) model}. While ShopMind generates behavior from mental states, MSR tackles the inverse problem of inferring latent mental states from observed behavior, supporting more accurate user understanding and behavior prediction on real-world data.

%Experimental results show that ShopMind outperforms existing simulators in behavioral realism and personalization. 

Experiments show that ShopMind achieves gains of 18.0\% in personalization and 19.6\% in behavioral realism over the strongest baselines. Moreover, the mental states reconstructed by MSR improve consumption value measurement and CTR prediction, indicating that the mental-state supervision produced by ShopMind facilitates the understanding and prediction of real user behavior.

In summary, the main contributions of this work can be summarized as follows:
%In summary, the key contributions of this work are as follows:
\begin{itemize}
    \item We propose \textbf{ShopMind}, a mental world model for e-commerce user simulation. ShopMind models consumers’ mental states and their evolution, capturing how users interpret shopping information and make decisions to enable personalized and realistic simulation.
    \item ShopMind synthesizes shopping trajectories paired with evolving mental states. A \textbf{Mental State Reconstruction (MSR) model} is trained on these data to recover latent mental states from observable behavior, supporting user understanding and behavior prediction.
    \item Experiments show that ShopMind improves personalization by 18.0\% and behavioral realism by 19.6\% over the strongest baselines, while MSR-reconstructed mental states significantly improve consumption value measurement and future behavior prediction.
    
    %Experimental results show that ShopMind outperforms existing simulators in behavioral realism and personalization, while MSR-reconstructed mental states significantly improve consumption value measurement and future behavior prediction.
\end{itemize}

\section{Related Work}
\vspace{-0.5em}

\noindent\textbf{E-commerce User Simulation.}
LLM-based virtual shoppers have recently emerged as a promising approach to e-commerce user simulation.
PAARS constructs persona-conditioned shopper populations from historical data to better match real user distributions~\citep{mansour2025paars}. 
\citet{lu2026multiturn} show that supervised fine-tuning on real shopping trajectories improves behavioral fidelity, while Shop-R1 further employs reinforcement learning to align generated actions with human behavior~\citep{zhang2026shopr1}. 
Customer-R1 extends this line toward personalized simulation by incorporating explicit user personas~\citep{wang2025customerr1}, and OPeRA and ShopperBench further emphasize persona-aware, multi-step shopping behavior~\citep{wang2026opera,ling2026shopperbench}. 
Despite these advances, existing approaches still predict behavior largely from observable signals, with limited consideration of how consumers interpret information and how their mental states evolve over time.

\noindent\textbf{World Model.}
World models learn environment dynamics to anticipate the outcomes of different actions and support planning~\citep{ha2018worldmodels}. 
Representative methods such as PlaNet, Dreamer, and MuZero use learned or imagined future trajectories to guide decision-making~\citep{hafner2019planet,hafner2020dreamer,schrittwieser2020muzero}, but primarily model the external world. Recent work extends this idea to the mental world by explicitly modeling latent human states, including beliefs, goals, and intentions, and how they evolve with observations and actions~\citep{liu2026mentalreview,fei2026mwm}. 
\citet{fei2026mwm} further formulate mental world modeling as jointly tracking physical and mental states to better predict human decisions. Building on this perspective, ShopMind brings mental world modeling to e-commerce user simulation by continuously tracking consumers' beliefs, uncertainties, and intentions and using their evolution to drive subsequent behavior.

\vspace{-1em}
\section{Problem Formulation}
\vspace{-0.5em}
We study personalized and realistic user behavior simulation in e-commerce. We formulate the task as follows. For each user $u$, let $P_u$ denote the user profile capturing relatively stable user characteristics, and $T_u$ the shopping task specifying the goal and constraints of the current episode. At step $t$, $O_t$ denotes the observable shopping context, while $H_t$ summarizes the interaction history and retrieved memory. The goal is to model a user policy $\pi$ that generates the next action $A_t$ conditioned on the available context, i.e., $A_t \sim \pi(\cdot \mid P_u, T_u, O_t, H_t)$. Rather than directly mapping observable context to actions, ShopMind models user behavior through an evolving mental state $M_t$, where $M_t = F_{\theta}(P_u, T_u, O_t, H_t)$ and $A_t \sim \pi_{\theta}(\cdot \mid M_t, P_u, T_u, O_t, H_t)$. Here, $F_{\theta}$ models mental-state formation and transition, while $M_t$ captures the user's current beliefs, uncertainties, and intentions. After executing $A_t$, the environment returns a new observation $O_{t+1}$, which further updates the mental state, forming the sequential process $O_t \rightarrow M_t \rightarrow A_t \rightarrow O_{t+1} \rightarrow M_{t+1}$.

\vspace{-0.5em}
\section{ShopMind: Mental World Model}
\vspace{-0.5em}

\begin{figure*}[t!]
    \centering    
    \includegraphics[width=\linewidth, keepaspectratio]{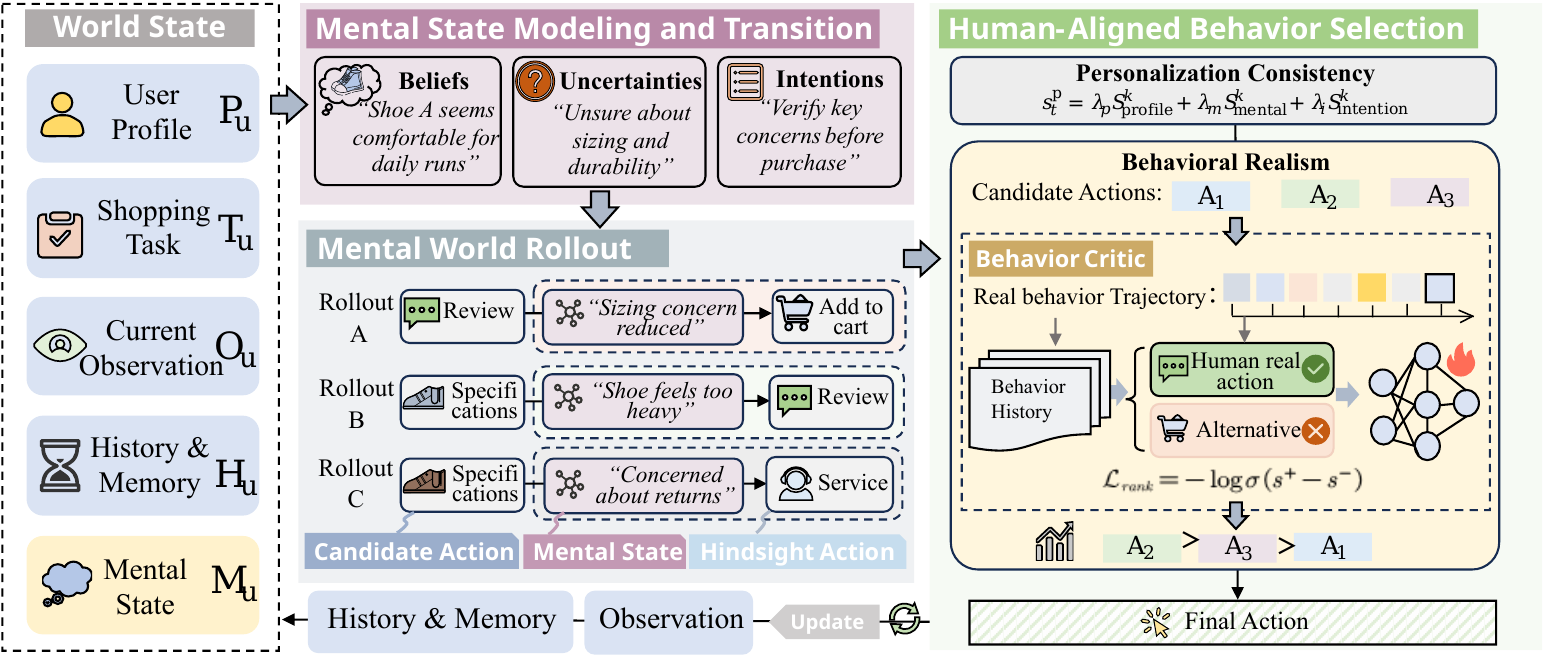}
    \vspace{-1em}
    \caption{Overview of the \textbf{ShopMind} framework. The model maintains an evolving mental state, performs mental-world rollouts over candidate behaviors, and selects the final action by jointly considering personalization consistency and behavioral realism through a learned Behavior Critic.
    }
    \label{fig:method}
    \vspace{-1em}
\end{figure*}

Existing methods typically predict user actions from observable shopping context, overlooking the evolving mental state that shapes behavior. ShopMind addresses this limitation by modeling a dynamic mental world that captures the consumer’s internal mental state, and uses it to guide subsequent behavior generation. As shown in Figure~\ref{fig:method}, the framework consists of three stages: \textbf{(1) Mental State Modeling and Transition} maintains a structured representation of the consumer’s mental state (beliefs, uncertainties, and intentions), and updates it as new information arrives; \textbf{(2) Mental-World Rollout} generates candidate actions and simulates their future environment, mental-state, and behavioral trajectories; and \textbf{(3) Human-Aligned Behavior Selection} ranks candidates by individual consistency and behavioral realism, combining LLM-based trajectory assessment with a Behavior Critic trained on real consumer trajectories. Together, these stages capture how evolving mental states shape behavior while producing personalized and realistic user behaviors.

%The framework consists of three stages: \textbf{(1) Mental State Modeling and Transition} constructs a structured representation of the consumer’s beliefs, attitudes, uncertainties, and intentions, and continuously updates it as new information is acquired; \textbf{(2) Mental-World Rollout} generates multiple candidate actions and simulates their possible future consequences, including subsequent environment changes, mental-state transitions, and behaviors; and \textbf{(3) Human-Aligned Behavior Selection} evaluates each candidate in terms of personalized consistency and behavioral realism, using an LLM-based trajectory assessment together with a Behavior Critic trained on real consumer trajectories to determine the final action. Together, these stages allow ShopMind to model not only what a consumer does, but also how their internal state evolves and shapes behavior, while keeping the generated trajectories both personalized and consistent with real consumer behavior.

\vspace{-0.5em}
\subsection{Mental State Modeling and Transition}
\vspace{-0.5em}
Theory of Mind~\citep{premack1978chimpanzee} suggests that human behavior depends on internal mental states, such as beliefs, uncertainties, and intentions, which evolve as new information is acquired. Inspired by this perspective, ShopMind represents the consumer’s mental state in a structured form, making its evolution explicit and consistently trackable throughout the shopping process. 

Guided by Theory of Mind, we represent the consumer's mental state at time step $t$ as
\[
M_t =
\left(
B_t,
U_t,
I_t,
E_t
\right),
\]
where $B_t$ summarizes the beliefs formed from observed information, including the user's understanding and judgments of the shopping context and candidate products; $U_t$ captures unresolved uncertainties that may influence subsequent decisions; $I_t$ specifies the user's current intention and immediate goal; and $E_t$ stores the supporting evidence for these mental states. 

Since the mental state depends on both the current observation and interaction history, ShopMind maintains a memory $H_t$ of decision-relevant past behaviors and observations. The mental state is initialized from the user profile, shopping task, and initial observation, and is subsequently updated using the previous state, interaction memory, and latest action and observation:
$$
M_0 =
F_{\theta}^{\mathrm{init}}(P_u,T_u,O_0),
\qquad
M_{t+1}
=
F_{\theta}(M_t,P_u,T_u,H_t,A_t,O_{t+1}).
$$
A key challenge in modeling mental-state dynamics is that users' mental states do not necessarily change with every behavioral  interaction.  Treating every low-level action as a mental-state transition would introduce spurious fluctuations and undermine the temporal coherence of the inferred mental trajectory. To address this, ShopMind introduces an update gate that triggers a state transition only when the newly observed information is sufficiently informative to update the current mental state; otherwise, $M_{t+1}=M_t$. This mechanism preserves stable mental states across behaviorally trivial interactions while allowing meaningful cognitive changes to emerge when new evidence is acquired.  

\subsection{Mental World Rollout}
Given the mental state, ShopMind first generates a set of candidate actions. However, evaluating candidates only at the current step can be myopic, as actions with similar immediate plausibility may lead to different future trajectories. For example, when a user is comparing two products, both examining the current product further and switching to another candidate may seem reasonable at the current step, yet lead to different trajectories: one may reduce uncertainty and clarify preference, while the other may provide little new information and prolong the comparison. Accordingly,  ShopMind rolls out each candidate independently, simulating subsequent environment changes, mental-state transitions, and behaviors, thereby promoting more coherent and realistic decision trajectories.

We define the current decision context as \(J_t=(P_u,T_u,O_t,H_t,M_t)\), consisting of the user profile, shopping task, current observation, interaction memory, and mental state. Given \(J_t\), ShopMind generates \(K\) candidate actions: \(\mathcal{C}_t=\{a_t^1,\ldots,a_t^K\}\). To ensure executability, candidate generation is constrained by the valid
action space under the environment state. Each candidate
\(a_t^k\) is then rolled out for \(L\) steps in an independent copy of the
environment, producing an imagined trajectory
\[
\hat{\tau}_t^k=
\left(
a_t^k,
\hat{O}_{t+1}^k,
\hat{M}_{t+1}^k,
\ldots,
\hat{O}_{t+L}^k,
\hat{M}_{t+L}^k
\right).
\]
The rollout jointly simulates how each candidate may reshape the environment
and the user's mental state, allowing the current action to be evaluated
through its possible future trajectory. 

%Rollouts are isolated across candidates, and imagined observations are never written back to the actual interaction history or mental state, preventing simulated futures from contaminating the ongoing interaction.

\subsection{Human-Aligned Behavior Selection}
For each candidate trajectory, ShopMind evaluates it along two dimensions: Personalized Consistency and Behavioral Realism. The former assesses alignment with the user’s profile and mental state. Behavioral Realism, in contrast, measures real-world plausibility, distinguishing among individually consistent actions and favoring those that better reflect real-world behavioral patterns.

\noindent\textbf{Personalized Consistency.}
It considers three aspects of the rollout trajectory: profile alignment $S_{\mathrm{profile}}^k$ with the user's characteristics, mental-state coherence $S_{\mathrm{mental}}^k$ throughout the rollout, and intention consistency $S_{\mathrm{intention}}^k$ between the current intention and subsequent behaviors.
An LLM-based evaluator scores these dimensions, and the resulting scores are aggregated as
\[
s_t^{k,\mathrm{pc}}
=
\lambda_p S_{\mathrm{profile}}^k
+
\lambda_m S_{\mathrm{mental}}^k
+
\lambda_i S_{\mathrm{intention}}^k,
\]
to quantify the consistency of the trajectory with the user's mental state and behavioral tendencies.

\noindent\textbf{Behavioral Realism.}
Multiple candidate actions may be similarly plausible for a given user, yet differ in behavioral realism. When uncertain about fit, for instance, the user may check the size guide, read reviews, or contact customer service, although these actions vary in how closely they reflect real-world behavioral patterns. Therefore, we train a \textbf{Behavior Critic} on real consumer trajectories to assign each candidate a scalar behavioral plausibility score, indicating how well the action matches real user behavior under the current shopping context: \(c_t^k = C_\phi(H_t, O_t, a_t^k)\), where \(H_t\) and \(O_t\) denote the interaction history and observation, respectively. Higher scores indicate greater behavioral realism. The critic is trained with a pairwise ranking objective, encouraging the observed next action \(a^+\) to score higher than an alternative \(a^-\): \( C_\phi(h_t,a^+) > C_\phi(h_t,a^-). \) We optimize \[ \mathcal{L}_{\mathrm{critic}} = -\log \sigma\!\left( C_\phi(h_t,a^+) - C_\phi(h_t,a^-) \right). \]
For final selection, we rank candidates by personalized consistency
\(s_t^{k,\mathrm{pc}}\) and define the top-two margin as
\(\Delta_t=s_t^{(1),\mathrm{pc}}-s_t^{(2),\mathrm{pc}}\).
If \(\Delta_t\geq\tau_{\mathrm{margin}}\), we select the top-ranked candidate
unless strongly rejected by the Behavior Critic.
Otherwise, candidates with
\(s_t^{k,\mathrm{pc}}\geq\tau_{\mathrm{compat}}\)
form \(\mathcal{C}_t^{\mathrm{pc}}\), from which the Behavior Critic selects
the most realistic action:
\[
A_t
=
\arg\max_{a_t^k\in\mathcal{C}_t^{\mathrm{pc}}}
C_\phi(H_t,O_t,a_t^k).
\]
This design preserves clear user-specific behavioral tendencies while using
real consumer behavior to resolve ambiguous choices, balancing personalization
and behavioral realism.

\section{Learning Mental State Reconstruction from ShopMind}

Beyond behavior simulation, ShopMind provides scalable supervision for reconstructing latent mental states from observed behavior. Real-world e-commerce logs capture user interaction histories, but rarely record how mental states evolve during shopping. These latent states provide a richer representation of users for understanding their values and predicting future behavior, yet obtaining such annotations through surveys or interviews is costly and difficult to scale. ShopMind instead jointly synthesizes behavioral trajectories and aligned mental-state transitions, enabling large-scale behavior--mental-state supervision. After quality filtering, we use these data to train a Mental State Reconstruction (MSR) model that infers latent mental states from real behavioral traces and supports downstream consumer value measurement and behavior prediction.

\subsection{Data Construction}

Given the lack of continuous mental-state annotations in real-world logs, we leverage ShopMind to synthesize complete shopping trajectories together with the mental-state sequences that evolve throughout the interaction. Each trajectory starts from the initial shopping context and ends with a purchase or session termination. We organize each trajectory as a single MSR training instance, pairing the complete observable interaction with its aligned mental-state sequence:
\[
\mathcal{D}_{\mathrm{MSR}}
=
\left\{
\left(
X^{(i)}_{1:T_i},
M^{(i)}_{1:K_i}
\right)
\right\}_{i=1}^{N},
\]
where $X^{(i)}_{1:T_i}$ denotes the user profile and interaction trajectory, and $M^{(i)}_{1:K_i}$ denotes the corresponding sequence of latent mental states, with each mental state potentially spanning multiple consecutive interactions. To ensure dataset quality, we first perform deterministic checks on purchase outcomes, action execution, and mental-state transitions, followed by an LLM-based evaluation of task fit, decision consistency, state--action consistency, and evidence quality. We further conduct human review to validate trajectory quality. After filtering, the final dataset contains 5K high-quality samples for MSR training. Quality-control results are reported in Appendix~\ref{app:msr_quality}.

\subsection{Mental State Reconstruction Model}
We train a Mental State Reconstruction (MSR) model on the quality-controlled trajectories to recover latent mental-state evolution from observed shopping behavior. Given an observable trajectory $X_{1:T}$, MSR reconstructs the corresponding mental-state sequence:
\(
\hat{M}_{1:T}=f_{\theta}(X_{1:T}),
\)
where each $\hat{M}_t=\{\hat{B}_t,\hat{U}_t,\hat{I}_t\}$ represents the reconstructed beliefs, uncertainties, and intentions at step $t$. We use Qwen3-8B~\citep{yang2025qwen3} as the backbone and fine-tune it on the 5K quality-filtered trajectories using supervised fine-tuning (SFT). MSR is formulated as a conditional generation task, producing the same structured mental-state representation as ShopMind. The training objective is
\[
\mathcal{L}_{\mathrm{MSR}}
=
-\sum_{i=1}^{N}
\log p_{\theta}
\left(
M^{(i)}_{1:T_i}
\mid
X^{(i)}_{1:T_i}
\right).
\]
This objective enables MSR to reconstruct how latent mental states evolve throughout an observed shopping trajectory. Additional training details are provided in Appendix~\ref{app:msr}.

\begin{figure*}[t!]
    \centering    
    \includegraphics[width=\linewidth, keepaspectratio]{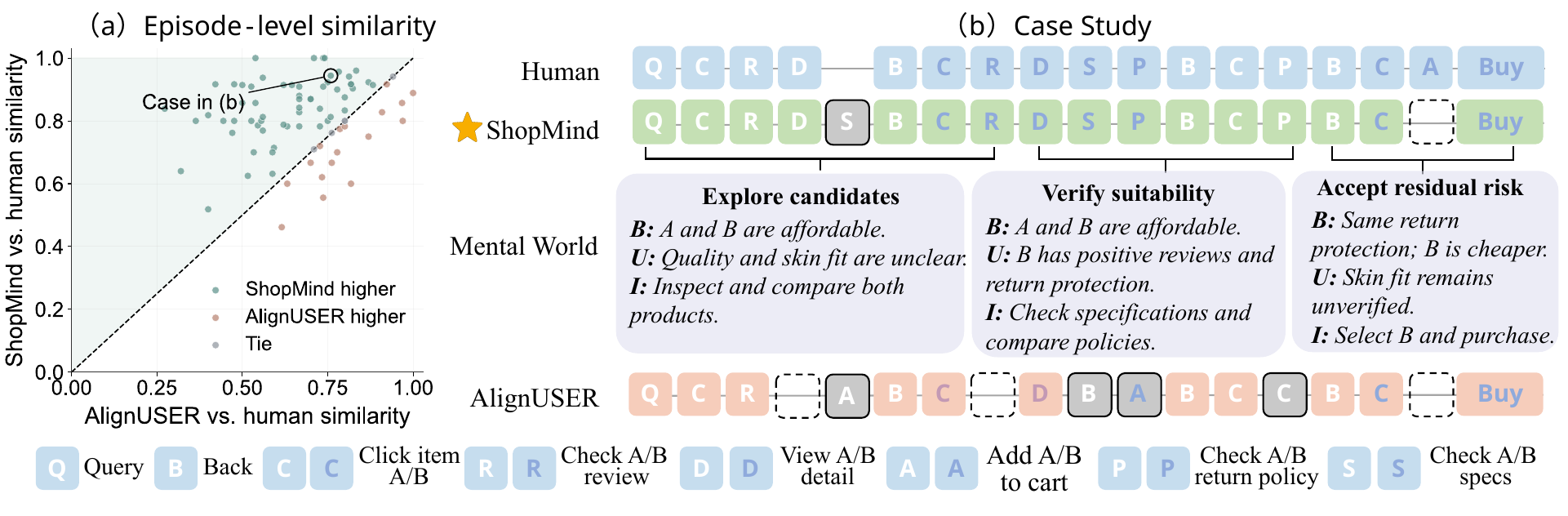}
    \vspace{-1em}
    \caption{Qualitative comparison of ShopMind and AlignUSER~\citep{bougie2026alignuser}, a strong baseline, on 80 randomly sampled episodes from the Taobao real log dataset. (a) ShopMind shows higher similarity to human trajectories. (b) A case illustrates how mental-world modeling helps ShopMind better follow human decision dynamics, while AlignUSER exhibits larger deviations. 
    }
    \label{fig:case}
    \vspace{-1em}
\end{figure*}

\section{Experiments}
Our experiments focus on two research questions: \textit{(1) RQ1: Can ShopMind generate more personalized and realistic shopping behaviors than existing e-commerce user simulators?} \textit{(2) RQ2: Can MSR trained on ShopMind-synthesized data improve real-user understanding and behavior prediction, demonstrating the real-world transferability of the synthesized mental-state supervision?} We further conduct ablation studies to quantify the contribution of each key component.

\subsection{Experimental Setup}

\noindent\textbf{Implementation Details.} ShopMind uses DeepSeek-V4-Pro~\citep{deepseekai2026deepseekv4} as its backbone, with \(K=4\) candidates and an \(L=2\) rollout horizon. The Behavior Critic is initialized from Qwen3-4B~\citep{yang2025qwen3} and fine-tuned with LoRA using LLaMA-Factory~\citep{zheng2024llamafactory} on two NVIDIA A100 80GB GPUs. More details are provided in Appendix~\ref{app:shopmind}.

\noindent\textbf{Datasets and Models.} For RQ1, we evaluate ShopMind on next-action prediction and multi-step trajectory reproduction using OPeRA~\citep{wang2026opera}, ShopCART~\citep{lu2026multiturn}, and Taobao real shopping logs. We compare against representative LLM-based user simulation methods, including SimUSER~\citep{bougie2025simuser}, DGDPO~\citep{liu2026dgdpo}, AlignUSER~\citep{bougie2026alignuser}, IToM~\citep{chen2026itom}, Shop-R1~\citep{zhang2026shopr1}, ContextSim~\citep{bougie2026contextsim}, and Customer-R1~\citep{wang2025customerr1} (see Appendix~\ref{app:comparison}). For RQ2, we evaluate MSR on two tasks: consumption value measurement and CTR prediction. On B2V-Bench~\citep{hou2026behavior2value}, we compare against 10 strong LLMs and B2V-Verifier to assess whether mental states improve value measurement. For CTR prediction, we construct a test set from real Taobao logs, using DeepSeek-V4-Pro~\citep{deepseekai2026deepseekv4} and Qwen2.5-72B~\citep{qwen2025qwen25} as backbones (see Appendix~\ref{app:downstream}).

%For RQ1, we evaluate ShopMind on next-action prediction and multi-step trajectory reproduction using OPeRA, ShopCART, and Taobao real-world shopping logs. We compare against representative e-commerce user simulators, including SimUSER, DGDPO, AlignUSER, and Customer-R1, retaining only baselines that can be fairly evaluated under matched inputs and action spaces on each dataset. ShopMind is built on DeepSeek-V4-Pro.

\noindent\textbf{Evaluation Metrics.} RQ1 is evaluated using Individual Action Distribution JS, User-level Spearman $\rho$, Persona Shuffle $\Delta$, and Group Gap Error for individual consistency; Exact Action Accuracy, Action Type Macro-F1, Top-$k$ Action Accuracy, Non-Terminal Action Accuracy, Purchase F1, and Exit F1 for local behavioral fidelity; and Action Transition JS, Session Length Error, Purchase Rate Error, and Action JS for sequential realism. Jensen--Shannon divergence is used for distributional comparison~\citep{lin1991divergence}. RQ2 adopts the evaluation protocol of B2V-Bench~\citep{hou2026behavior2value} for consumption value measurement, and uses AUC, GAUC, and PR-AUC for CTR prediction.

\begin{table}[t]
\centering
\caption{
Performance comparison of individual consistency.
ShopMind achieves the best performance across all metrics. $\star$, $\dagger$, and $\ddagger$ denote the first-, second-, and third-best results, respectively.
}
\label{tab:rq1_individual}
\fontsize{8pt}{8.5pt}\selectfont
\setlength{\tabcolsep}{3.2pt}

\resizebox{\linewidth}{!}{
\begin{tabular}{ll|cccc|cccc}
\toprule
& & \multicolumn{4}{c}{\textbf{OPeRA}}
& \multicolumn{4}{c}{\textbf{Taobao}} \\
\cmidrule(lr){3-6}
\cmidrule(lr){7-10}

\textbf{Method}
& \textbf{Method Type}
& \shortstack{\textbf{Ind.}\\\textbf{JS} $\downarrow$}
& \shortstack{\textbf{Spearman}\\$\uparrow$}
& \shortstack{\textbf{Gap}\\\textbf{Err.} $\downarrow$}
& \shortstack{\textbf{$\Delta$}\\\textbf{JS} $\uparrow$}
& \shortstack{\textbf{Ind.}\\\textbf{JS} $\downarrow$}
& \shortstack{\textbf{Spearman}\\$\uparrow$}
& \shortstack{\textbf{Gap}\\\textbf{Err.} $\downarrow$}
& \shortstack{\textbf{$\Delta$}\\\textbf{JS} $\uparrow$} \\
\midrule

Zero-shot
& Prompting
& 31.02
& 0.334
& 0.054
& $-0.44$
& 34.28
& 0.281
& 0.061
& $-0.62$ \\

SFT
& SFT
& 26.48
& 0.330
& 0.051
& $-1.41$
& 29.82
& 0.308
& 0.051
& $-1.94$ \\

SFT+Rationale
& SFT
& 24.31
& 0.363
& 0.039
& $-2.61$
& \third{25.16}
& 0.337
& \third{0.042}
& $-0.91$ \\

SimUSER v2
& Agent
& 32.24
& 0.238
& 0.055
& $+0.29$
& 31.82
& 0.254
& 0.057
& $+1.14$ \\

DGDPO
& Agent
& 30.22
& 0.296
& 0.037
& \third{$+1.74$}
& 29.74
& 0.303
& 0.044
& $+1.86$ \\

ContextSim
& Agent
& 32.04
& 0.214
& 0.054
& \second{$+2.44$}
& 28.96
& 0.319
& 0.049
& \third{$+2.17$} \\

IToM
& Agent
& 30.25
& 0.315
& 0.054
& $+1.25$
& 27.83
& 0.326
& 0.047
& \second{$+2.35$} \\

AlignUSER
& Agent
& \second{20.36}
& \second{0.376}
& \second{0.035}
& $-0.36$
& \second{23.47}
& \third{0.348}
& \second{0.039}
& $+0.18$ \\

Shop-R1
& RL
& \third{22.60}
& \third{0.368}
& \third{0.036}
& $-1.27$
& 25.72
& 0.341
& 0.045
& $-1.52$ \\

Customer-R1
& RL
& 27.46
& 0.347
& 0.043
& $-1.53$
& 26.41
& \second{0.352}
& 0.043
& $-0.94$ \\

\midrule

\textbf{ShopMind}
& \textbf{Mental World}
& \first{17.21}
& \first{0.438}
& \first{0.028}
& \first{$\mathbf{+4.12}$}
& \first{18.63}
& \first{0.421}
& \first{0.031}
& \first{$\mathbf{+4.36}$} \\

\bottomrule
\end{tabular}
}
\vspace{-2em}
\end{table}

\begin{table}[t]
\centering
\caption{
Performance comparison of local behavior prediction.
ShopMind achieves the best performance on most metrics. $\star$, $\dagger$, and $\ddagger$ denote the first-, second-, and third-best results, respectively.
}
\label{tab:rq1_local}

\fontsize{7pt}{6.8pt}\selectfont
\setlength{\tabcolsep}{3.8pt}

\resizebox{\linewidth}{!}{
\begin{tabular}{lccccc|ccccc}
\toprule
& \multicolumn{5}{c}{\textbf{OPeRA}}
& \multicolumn{5}{c}{\textbf{ShopCART}} \\
\cmidrule(lr){2-6}
\cmidrule(lr){7-11}

\textbf{Method}
& \shortstack{\textbf{Exact}\\\textbf{Acc.} $\uparrow$}
& \shortstack{\textbf{Non-Term.}\\\textbf{Acc.} $\uparrow$}
& \shortstack{\textbf{Macro}\\\textbf{F1} $\uparrow$}
& \shortstack{\textbf{Top-3}\\\textbf{Acc.} $\uparrow$}
& \shortstack{\textbf{Buy}\\\textbf{F1} $\uparrow$}
& \shortstack{\textbf{Exact}\\\textbf{Acc.} $\uparrow$}
& \shortstack{\textbf{Macro}\\\textbf{F1} $\uparrow$}
& \shortstack{\textbf{Top-3}\\\textbf{Acc.} $\uparrow$}
& \shortstack{\textbf{Buy}\\\textbf{F1} $\uparrow$}
& \shortstack{\textbf{Exit}\\\textbf{F1} $\uparrow$} \\
\midrule

Zero-shot
& 6.10
& 4.85
& 32.50
& 8.98
& \second{67.31}
& 6.96
& 17.60
& 9.23
& 36.76
& 0.62 \\

SFT
& 29.38
& 27.50
& 45.45
& 42.35
& 20.69
& 20.83
& 40.62
& 41.28
& 35.46
& 33.71 
 \\

SFT+Rationale
& 33.04
& 29.89
& 42.32
& 46.23
& 52.05
& 15.14
& 39.09
& 43.60
& \second{59.70}
& 14.25 \\

SimUSER v2
& 5.32
& 3.55
& 35.81
& 12.31
& 42.67
& 4.26
& 27.98
& 1.23
& 29.51
& 1.23 \\

DGDPO
& 5.99
& 5.14
& 34.33
& 9.20
& \second{67.31}
& 6.41
& 20.72
& 10.85
& 41.63
& 3.17 \\

ContextSim
& 16.19
& 12.96
& 35.46
& 28.49
& 42.67
& 11.05
& 35.79
& 35.51
& 20.93
& \third{36.45} \\

IToM
& 6.21
& 4.70
& 31.21
& 8.76
& \first{68.63}
& 7.57
& 18.08
& 9.49
& 37.16
& 2.45 \\

AlignUSER
& \third{35.92}
& \second{33.62}
& \third{47.81}
& \second{47.89}
& 51.85
& \third{22.98}
& \second{48.40}
& \third{44.13}
& 31.37
& \second{58.77}\\

Shop-R1
& \second{37.18}
& \third{32.95}
& \second{49.26}
& \third{47.31}
& 52.74
& 16.45
& \third{41.49}
& \second{44.47}
& \first{60.61}
& 22.83 \\

Customer-R1
& 27.82
& 24.63
& 18.74
& 31.86
& 0.00
& \second{28.02}
& 15.11
& 30.46
& 0.00
& \first{94.85} \\

\midrule

\textbf{ShopMind}
& \first{41.22}
& \first{37.10}
& \first{52.89}
& \first{60.85}
& \third{54.23}
& \first{31.29}
& \first{49.14}
& \first{45.94}
& \third{58.53}
& 27.45 \\

\bottomrule
\end{tabular}
}
\vspace{-1em}
\end{table}

%-------------------------实验二
\begin{table}[t]
\centering
\caption{
Performance comparison of long-horizon behavioral realism. ShopMind achieves consistent gains across all metrics. $\star$, $\dagger$, and $\ddagger$ denote the first-, second-, and third-best results, respectively.
}
\label{tab:rq1_long}
\fontsize{8.5pt}{9pt}\selectfont
\setlength{\tabcolsep}{3.2pt}

\resizebox{\textwidth}{!}{
\begin{tabular}{lcccc|cccc|cccc}
\toprule
& \multicolumn{4}{c}{\textbf{OPeRA}}
& \multicolumn{4}{c}{\textbf{ShopCART}}
& \multicolumn{4}{c}{\textbf{Taobao}} \\
\cmidrule(lr){2-5}
\cmidrule(lr){6-9}
\cmidrule(lr){10-13}

\textbf{Method}
& \shortstack{\textbf{Trans.}\\\textbf{JS} $\downarrow$}
& \shortstack{\textbf{Len.}\\\textbf{Err.} $\downarrow$}
& \shortstack{\textbf{Buy}\\\textbf{Err.} $\downarrow$}
& \shortstack{\textbf{Action}\\\textbf{JS} $\downarrow$}
& \shortstack{\textbf{Trans.}\\\textbf{JS} $\downarrow$}
& \shortstack{\textbf{Len.}\\\textbf{Err.} $\downarrow$}
& \shortstack{\textbf{Buy}\\\textbf{Err.} $\downarrow$}
& \shortstack{\textbf{Action}\\\textbf{JS} $\downarrow$}
& \shortstack{\textbf{Trans.}\\\textbf{JS} $\downarrow$}
& \shortstack{\textbf{Len.}\\\textbf{Err.} $\downarrow$}
& \shortstack{\textbf{Buy}\\\textbf{Err.} $\downarrow$}
& \shortstack{\textbf{Action}\\\textbf{JS} $\downarrow$} \\
\midrule

Zero-shot
& 38.06
& 7.47
& 31.11
& 23.53
& 51.73
& 2.80
& 48.18
& 31.20
& 44.82
& 6.95
& 27.64
& 25.81 \\

SFT
& 18.05
& 4.78
& 43.33
& 14.65
& 23.84
& 2.64
& 6.37
& 12.76
& \third{17.83}
& 4.35
& 12.77
& 13.26 \\

SFT+Rationale
& \third{14.45}
& 5.13
& 15.56
& \second{10.84}
& 33.05
& 4.27
& \third{1.40}
& 15.30
& 20.31
& 4.88
& 10.42
& 13.95 \\

SimUSER v2
& 32.74
& 7.21
& 12.22
& 25.04
& 28.07
& 3.62
& 19.89
& 21.78
& 29.63
& 6.38
& 14.21
& 22.16 \\

DGDPO
& 39.14
& 7.11
& 34.44
& 22.16
& 41.36
& 3.18
& 31.27
& 22.43
& 34.88
& 6.17
& 24.93
& 20.42 \\

ContextSim
& 17.56
& 7.20
& 12.22
& 18.32
& \third{20.95}
& \third{2.38}
& 8.40
& \third{11.59}
& 18.94
& 5.22
& 9.18
& 14.03 \\

IToM
& 37.85
& 7.54
& 34.44
& 21.43
& 48.08
& 2.78
& 43.42
& 26.00
& 40.26
& 6.84
& 29.77
& 23.65 \\

AlignUSER
& \second{13.14}
& \second{4.37}
& \second{4.44}
& \third{11.38}
& \second{19.73}
& 2.41
& \second{0.28}
& \second{6.18}
& \second{16.72}
& \third{4.12}
& \third{5.36}
& \third{12.47}\\

Shop-R1
& 15.82
& 5.06
& \third{9.74}
& 12.63
& 32.25
& 4.14
& 1.68
& 13.82
& 27.14
& 4.56
& \second{3.92}
& \second{11.84} \\

Customer-R1
& 25.68
& \third{4.61}
& 14.83
& 16.27
& 50.66
& \second{2.21}
& 9.80
& 17.56
& 31.77
& \second{3.91}
& 7.13
& 14.28 \\

\midrule

\textbf{ShopMind}
& \first{10.92}
& \first{3.86}
& \first{3.71}
& \first{8.93}
& \first{16.84}
& \first{1.98}
& \first{0.21}
& \first{5.24}
& \first{12.18}
& \first{3.62}
& \first{2.84}
& \first{9.41} \\

\bottomrule
\end{tabular}
}
\vspace{-2em}
\end{table}

\begin{table}[t]
\centering
\caption{
Consumption value measurement results. Best, second, and third results are marked by $\star$, $\dagger$, and $\ddagger$. MSR-reconstructed mental states provide the strongest gains for value measurement.
}
\label{tab:value_measurement}
\fontsize{14pt}{17pt}\selectfont
\setlength{\tabcolsep}{4.5pt}

\resizebox{\linewidth}{!}{
\begin{tabular}{llccc|cc}
\toprule
\textbf{Type}
& \textbf{Model}
& \shortstack{\textbf{NDCG}\\\textbf{@1} $\uparrow$}
& \shortstack{\textbf{NDCG}\\\textbf{@2} $\uparrow$}
& \shortstack{\textbf{NDCG}\\\textbf{@3} $\uparrow$}
& \shortstack{\textbf{Primary}\\\textbf{Acc.} $\uparrow$}
& \shortstack{\textbf{Primary Top-2}\\\textbf{Acc.} $\uparrow$} \\
\midrule

\multirow{8}{*}{General LLM}
& Claude-Haiku-4~\citep{anthropic2025claudehaiku}
& 0.7135 & 0.6721 & 0.6716 & 0.5743 & 0.7434 \\

& DeepSeek-V3.2~\citep{deepseekai2025deepseekv32}
& 0.7211 & 0.6967 & 0.6917 & 0.5613 & 0.7565 \\

& Gemini-2.5-Flash~\citep{google2025gemini25}
& 0.7750 & 0.7026 & 0.6746 & 0.5780 & 0.7174 \\

& Gemini-2.5-Pro~\citep{google2025gemini25}
& 0.5353 & 0.5170 & 0.5142 & 0.4442 & 0.5576 \\

& GLM-5~\citep{glm5team2026glm5}
& 0.7044 & 0.6729 & 0.6594 & 0.5780 & 0.7137 \\

& GPT-4o~\citep{openai2024gpt4o}
& 0.7211 & 0.6417 & 0.6345 & 0.5260 & 0.6449 \\

& GPT-5~\citep{openai2025gpt5}
& 0.6858 & 0.6873 & 0.7052 & 0.5483 & 0.7590 \\

& Qwen3-235B~\citep{yang2025qwen3}
& 0.7304 & 0.6998 & 0.6975 & 0.5743 & 0.7416 \\

\midrule

\multirow{3}{*}{Task-specific}
& B2V-Verifier~\cite{hou2026behavior2value}
& \second{0.8236}
& \second{0.8069}
& \second{0.7977}
& \second{0.6518}
& \second{0.8420} \\

& B2V + Zero-shot Mental State
& \third{0.7931}
& \third{0.7662}
& \third{0.7849}
& \third{0.6123}
& \third{0.7943} \\

\rowcolor{shopmindbg}
& \textbf{B2V + MSR Mental State (Ours)}
& \first{0.8474}
& \first{0.8714}
& \first{0.8590}
& \first{0.7868}
& \first{0.9154} \\

& $\Delta$ vs. B2V-Verifier
& \textbf{+2.89\%}
& \textbf{+7.99\%}
& \textbf{+7.68\%}
& \textbf{+20.71\%}
& \textbf{+8.72\%} \\

\bottomrule
\end{tabular}
}
\vspace{-1em}
\end{table}

\subsection{RQ1: Personalization and Behavioral Realism}
ShopMind is evaluated along two dimensions: \textbf{Individual Consistency} and \textbf{Behavioral Realism}. Behavioral realism is further examined at both local decision and long-horizon trajectory levels. Figure~\ref{fig:case} qualitatively illustrates closer alignment between ShopMind and human trajectories.

\noindent\textbf{Individual Consistency.}
We assess individual consistency by comparing simulated behaviors with real user behavior patterns.  Given each user's profile and shopping task, the model generates shopping trajectories, from which these distributions are computed. As shown in Table~\ref{tab:rq1_individual}, ShopMind achieves an average 18.0\% improvement in personalization, as measured by Individual JS across OPeRA and Taobao. We also perform a persona-shuffle test by randomly permuting user profiles while keeping all other inputs unchanged. ShopMind exhibits the largest $\Delta$JS, indicating a stronger dependence on user-specific information and greater sensitivity to individual differences.

%Table~\ref{tab:rq1_individual} shows that ShopMind achieves an average 18.0\% improvement in personalization, measured by Individual JS across OPeRA and Taobao. Importantly, ShopMind exhibits the largest increase in Individual JS after persona shuffling ($\Delta$JS), indicating that it captures meaningful user-specific behavioral differences beyond population-level patterns.

%Together with higher Spearman correlation and lower Group Gap Error, this indicates that ShopMind captures meaningful user-specific behavioral differences beyond population-level patterns.

\noindent\textbf{Behavioral Realism.}
We first evaluate next-action prediction given the user profile, current observation, and interaction history. As shown in Table~\ref{tab:rq1_local}, ShopMind achieves the best performance on most metrics across both datasets. Although some baselines achieve higher F1 on individual terminal actions, their predictions are often biased toward a specific outcome (e.g., Customer-R1 attains high Exit F1 but zero Buy F1). Furthermore, we evaluate behavioral realism over complete shopping sessions. Starting only from the initial user information and environment, each model acts autonomously until termination. Table~\ref{tab:rq1_long} shows that ShopMind achieves the best performance on all 12 metrics. Using Transition JS as the representative metric of behavioral realism, ShopMind achieves an average relative improvement of 19.6\% over the strongest baselines. These results suggest that modeling evolving mental world and future action consequences helps distinguish among plausible actions, improving both local prediction accuracy and long-horizon behavioral realism.

\begin{figure*}[t!]
    \centering    
    \includegraphics[width=\linewidth, keepaspectratio]{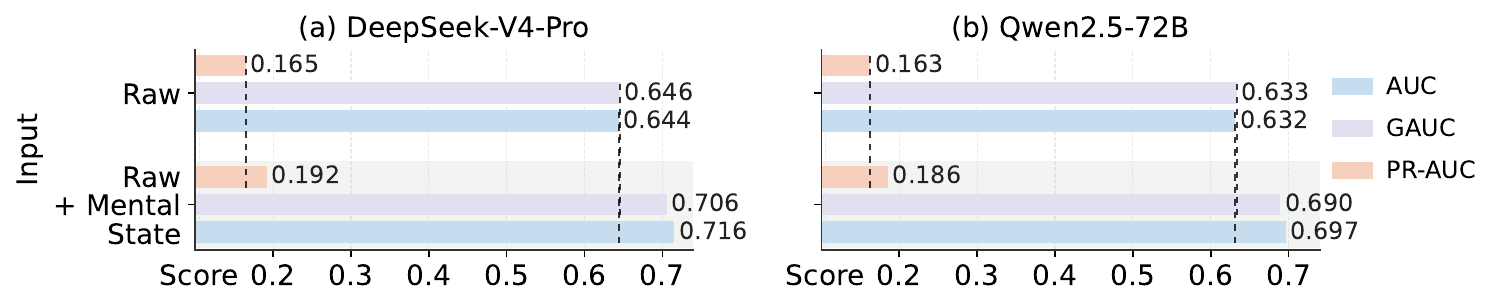}
    \vspace{-1.5em}
    \caption{CTR prediction performance with and without MSR-reconstructed mental states. Adding mental states consistently improves AUC, GAUC, and PR-AUC across both models.  
    }
    \label{fig:ctr}
    \vspace{-1em}
\end{figure*}

%----------------------------消融-------------------
\begin{table*}[t]
\centering
\caption{
Ablation study of ShopMind on Taobao logs across behavioral fidelity and personalization.
}
\label{tab:ablation}
\small
\setlength{\tabcolsep}{3.8pt}

\resizebox{\textwidth}{!}{
\begin{tabular}{lccc|cccc|ccc}
\toprule
& \multicolumn{3}{c}{\textbf{Local Fidelity}}
& \multicolumn{4}{c}{\textbf{Sequential Realism}}
& \multicolumn{3}{c}{\textbf{Individual Consistency}} \\
\cmidrule(lr){2-4}
\cmidrule(lr){5-8}
\cmidrule(lr){9-11}

\textbf{Variant}
& \shortstack{\textbf{Exact}\\\textbf{Acc.} $\uparrow$}
& \shortstack{\textbf{Macro}\\\textbf{F1} $\uparrow$}
& \shortstack{\textbf{Top-3}\\\textbf{Acc.} $\uparrow$}
& \shortstack{\textbf{Trans.}\\\textbf{JS} $\downarrow$}
& \shortstack{\textbf{Len.}\\\textbf{Err.} $\downarrow$}
& \shortstack{\textbf{Buy}\\\textbf{Err.} $\downarrow$}
& \shortstack{\textbf{Action}\\\textbf{JS} $\downarrow$}
& \shortstack{\textbf{Indiv.}\\\textbf{JS} $\downarrow$}
& \shortstack{\textbf{Spearman}\\$\rho$ $\uparrow$}
& \shortstack{\textbf{$\Delta$}\\\textbf{JS} $\uparrow$} \\
\midrule

%\rowcolor{gray!15}
\rowcolor{shopmindbg}
\textbf{ShopMind}
& \textbf{53.58} & \textbf{59.02} & \textbf{72.4}
& \textbf{12.18} & \textbf{3.62} & \textbf{2.84} & \textbf{9.41}
& \textbf{18.63} & \textbf{0.421} & \textbf{+4.36} \\

w/o Mental World
& 13.27 & 38.61 & 21.9
& 30.84 & 6.37 & 11.26 & 16.92
& 28.41 & 0.301 & +0.84 \\

w/o Rollout
& 50.84 & 56.47 & 68.9
& 22.63 & 5.41 & 7.38 & 15.27
& 20.12 & 0.397 & +3.72 \\

w/o Individual Consistency
& 50.96 & 55.83 & 68.2
& 13.86 & 3.91 & 3.31 & 10.24
& 29.16 & 0.318 & +1.43 \\

w/o Behavior Critic
& 44.72 & 49.86 & 62.5
& 18.41 & 4.67 & 6.14 & 14.38
& 20.37 & 0.389 & +3.58 \\
\bottomrule
\end{tabular}
}
\vspace{-1em}
\end{table*}

\vspace{-0.5em}
\subsection{RQ2: Real-World Utility of MSR}
\vspace{-0.5em}
We use consumption value measurement to test whether MSR provides a deeper understanding of users, and CTR prediction to assess its utility for subsequent behavior prediction.

\noindent\textbf{Consumption Value Measurement.}
For each real shopping trajectory, MSR reconstructs the latent mental state from observed behavior, which is integrated with the original behavior to predict the user's consumption value constructs. As shown in Table~\ref{tab:value_measurement}, B2V + MSR Mental State consistently outperforms both general LLMs and B2V-Verifier, with the largest gain on Primary Accuracy. In contrast, B2V + Zero-shot Mental State performs worse than B2V-Verifier, indicating that the gains stem from accurate mental-state reconstruction rather than simply adding inferred mental information, and further validating the effectiveness of the ShopMind-synthesized training data.

\noindent\textbf{CTR Prediction.} In this task, reconstructed mental states provide additional user-side signals for predicting subsequent click behavior. As shown in Figure~\ref{fig:ctr}, incorporating MSR-reconstructed mental states consistently improves performance across both backbones and all metrics, with full results reported in Appendix~\ref{app:ctr}. Directly using zero-shot mental states, however, degrades performance relative to the raw-behavior input. This comparison provides stronger evidence that the gains stem from ShopMind-synthesized mental-state supervision rather than from simply adding mental-state descriptions, demonstrating its transferability to real-world behavioral forecasting.

\vspace{-0.5em}
\subsection{Ablation Study}
\vspace{-0.5em}

We ablate each core component of ShopMind to examine its contribution. The ablation results in Table~\ref{tab:ablation} show that each component contributes to a distinct aspect of ShopMind. Removing the Mental World causes the largest degradation, highlighting its central role in maintaining coherent and personalized behavior. Without Rollout, performance declines mainly in sequential realism, suggesting that short-horizon simulation is important for maintaining realistic long-term trajectories. In contrast, ablating Individual Consistency primarily degrades user-level consistency and persona sensitivity, confirming its role in preserving individualized behavioral patterns. The Behavior Critic mainly affects local and sequential realism by filtering behaviorally implausible actions. Overall, all components contribute effectively to ShopMind's performance. Additionally, the hyperparameter sensitivity analysis in Appendix~\ref{app:hyperparameter} shows that ShopMind remains robust across a range of hyperparameter settings, with $K=4$ and $L=2$ achieving the best overall performance.

\vspace{-1em}
\section{Conclusion}
\vspace{-0.5em}
In this work, we introduce \textbf{ShopMind}, a \textbf{Mental World Model} for e-commerce user simulation that captures how consumers interpret shopping information and continuously form and update their beliefs, uncertainties, and intentions to drive more personalized and realistic behavior. Using the synthesized behavior--mental-state pairs, we further train a \textbf{Mental State Reconstruction model} that infers latent mental states from observed behaviors for real-user understanding and behavior prediction. Experiments show that ShopMind improves personalization and behavioral realism by 18\% and 19.6\%, respectively, while MSR-reconstructed mental states further enhance consumption value measurement and future behavior prediction, validating the utility of ShopMind-synthesized mental-state supervision. This work takes a step toward establishing Mental World Models as a general framework for modeling human cognition and behavior, with broader implications for recommender systems, user modeling, and computational cognitive science.
%This work lays the foundation for Mental World Models as a general framework for modeling human cognition and behavior, with potential implications for recommender systems, intelligent user modeling, and computational cognitive science.

\subsection*{AI use statement}

We used generative AI tools to assist with the generation of synthetic behavior--mental-state data and to polish the writing of the manuscript. All AI-assisted data and text were reviewed and verified by the authors. We take full responsibility for the final content of this work.

\subsection*{Ethics statement}

This work uses real-world user interaction data that were anonymized and de-identified prior to analysis. No personally identifiable information is included in the data used for our experiments, and all analyses are conducted solely on anonymized behavioral records for research purposes. Accordingly, the study is designed to preserve user privacy and minimize potential privacy risks.

\subsection*{Reproducibility statement}

The code and implementation details for this work are available in our anonymous repository at \url{https://anonymous.4open.science/r/ShopMind-0E94}. The main paper describes the model architecture, training procedure, datasets, baselines, and experimental settings, while additional implementation details and hyperparameter configurations are provided in the appendix.

\bibliography{iclr2027_conference}
\bibliographystyle{iclr2027_conference}

\appendix

\section*{Appendix}

\section{ShopMind Implementation Details}
\label{app:shopmind}

This section provides additional implementation details of mental world rollout, personalized
consistency evaluation, Behavior Critic training, and human-aligned behavior
selection.

% ============================================================
\subsection{Mental World Rollout}
\label{app:mental_rollout}

At each interaction step, ShopMind first constructs the current decision
context
\(
J_t=(P_u,T_u,O_t,H_t,M_t)
\)
and generates $K$ candidate actions
\(
\mathcal{C}_t
=
\{a_t^1,\ldots,a_t^K\}.
\)
We use $K=4$ in the default configuration. Candidate generation is constrained by the valid action space of the current
environment. Actions requiring an explicit target must refer to an available
product, option, or interface element. The candidate set is encouraged to
cover different plausible directions of the current decision, such as
acquiring additional information, comparing alternatives, progressing toward
purchase, or terminating the session when appropriate.

For each candidate $a_t^k$, ShopMind creates an independent copy of the
current environment and mental state and performs a counterfactual rollout of
$L$ steps. We use $L=2$ by default. The resulting imagined trajectory is
\[
\hat{\tau}_t^k=
\left(
a_t^k,
\hat{O}_{t+1}^k,
\hat{M}_{t+1}^k,
\ldots,
\hat{O}_{t+L}^k,
\hat{M}_{t+L}^k
\right).
\]
At every rollout step, the imagined action first changes the copied
environment. ShopMind then applies the same mental-state update mechanism to
determine whether the new observation should change the imagined mental
state. Subsequent behavior is generated conditioned on the resulting
environment and mental state, allowing the initial candidate to be evaluated
through the future trajectory it may induce. Candidate rollouts are completely isolated from one another. Environment
changes or imagined mental states from one branch cannot modify another
branch or the real interaction state. A rollout terminates when the maximum
horizon is reached, the simulated environment terminates, or no valid
continuation is available. Algorithm~\ref{alg:rollout} summarizes the rollout procedure.

\begin{algorithm}[t]
\caption{Mental World Rollout}
\label{alg:rollout}
\begin{algorithmic}[1]
\REQUIRE Decision context $J_t$, candidate number $K$, rollout horizon $L$
\STATE Generate valid candidates
$\mathcal{C}_t=\{a_t^1,\ldots,a_t^K\}$
\FOR{each candidate $a_t^k\in\mathcal{C}_t$}
    \STATE Copy the current environment and mental state
    \STATE Set the first imagined action to $a_t^k$
    \FOR{$l=0,\ldots,L-1$}
        \STATE Execute the imagined action in the copied environment
        \STATE Obtain the imagined observation $\hat{O}_{t+l+1}^k$
        \STATE Apply the mental-state update gate
        \IF{a mental-state update is triggered}
            \STATE Generate $\hat{M}_{t+l+1}^k$
        \ELSE
            \STATE Retain the preceding imagined mental state
        \ENDIF
        \IF{the environment terminates or $l=L-1$}
            \STATE \textbf{break}
        \ENDIF
        \STATE Generate the next valid rollout action
    \ENDFOR
\ENDFOR
\RETURN $\{\hat{\tau}_t^k\}_{k=1}^{K}$
\end{algorithmic}
\end{algorithm}

All main experiments use $K=4$ candidate actions and a rollout horizon of
$L=2$ unless otherwise specified. The sensitivity to $K$ and $L$ is
analyzed separately in Appendix~\ref{app:hyperparameter}.

% ============================================================
\subsection{Personalized Consistency Evaluation}
\label{app:personalized_consistency}

For each imagined trajectory, ShopMind evaluates whether the resulting
behavior remains consistent with the characteristics and current decision
state of the specific user. Given
\[
(P_u,T_u,M_t,H_t,\hat{\tau}_t^k),
\]
an LLM-based evaluator assesses three complementary dimensions:
profile alignment, mental-state coherence, and intention consistency.

\paragraph{Profile alignment.}
$S_{\mathrm{profile}}^k$ measures whether the imagined trajectory is
consistent with the user's relatively stable characteristics, preferences,
and behavioral tendencies. Explicit shopping constraints, such as budget and
product requirements, are also considered when judging whether a trajectory
is compatible with the user.

\paragraph{Mental-state coherence.}
$S_{\mathrm{mental}}^k$ measures whether the behaviors and mental-state
transitions throughout the rollout remain coherent with the user's current
beliefs and unresolved uncertainties. The evaluator considers whether each
simulated transition can be reasonably supported by the information available
along the imagined trajectory and whether the resulting mental-state evolution
remains internally consistent.

\paragraph{Intention consistency.}
$S_{\mathrm{intention}}^k$ measures whether the imagined behaviors are
consistent with the user's current high-level intention and whether they
meaningfully advance that intention. Since one intention may govern multiple
low-level actions, exact action--intention matching is not required. For
example, an intention to compare two shortlisted products may be realized
through several information-seeking and navigation actions.

The three scores are aggregated as
\[
s_t^{k,\mathrm{pc}}
=
\lambda_p S_{\mathrm{profile}}^k
+
\lambda_m S_{\mathrm{mental}}^k
+
\lambda_i S_{\mathrm{intention}}^k,
\]
where we set
\[
\lambda_p=0.35,\qquad
\lambda_m=0.30,\qquad
\lambda_i=0.35.
\]

% ============================================================
\subsection{Behavior Critic}
\label{app:behavior_critic}

Behavioral regularities can differ substantially across e-commerce
environments. Different datasets may have different action spaces, interface
structures, product exposure mechanisms, session patterns, and
population-level action distributions. We therefore train a separate
Behavior Critic for each evaluation dataset using its corresponding training
interactions.

Specifically, experiments on OPeRA use a Critic trained on OPeRA behavioral
trajectories, while ShopCART and Taobao use Critics trained on their respective
training data. All dataset-specific Critics share the same backbone,
fine-tuning strategy, pairwise training objective, and optimization
configuration. No interactions from an evaluation test split are used to
train its corresponding Critic.

\paragraph{Training data construction.}
For each real behavioral context $h_t$, the observed next action is treated
as the positive example $a^+$. We construct alternative actions $a^-$ under
the same interaction prefix so that the Critic learns to distinguish the
behavior actually observed in real trajectories from contextually mismatched
alternatives. Four types of negative samples are used. The first keeps the observed action
type but replaces its target with an incorrect product or SKU. The second
uses a different plausible action on the same target. The third involves
another product from the same category. The fourth uses a more broadly
mismatched random action. Together, these negatives expose the Critic to both
hard context-dependent alternatives and easier mismatched behaviors.

For each dataset-specific Critic, we sample $30{,}000$ training contexts, $5{,}000$ validation contexts, and $5{,}000$ test contexts from the corresponding dataset. Each context is paired with one observed positive action and four negative alternatives, yielding $120{,}000$, $20{,}000$, and $20{,}000$ positive--negative pairs for training, validation, and testing, respectively.

\paragraph{Model and training.}
We instantiate the Behavior Critic with
\texttt{Qwen3-4B-Instruct-2507} and train it as a reward model with a learned
scalar value head. LoRA is applied to all target modules with rank $16$,
scaling factor $32$, and dropout $0.05$. Each dataset-specific Critic is trained for three epochs with a learning rate
of $1\times10^{-4}$. We use cosine learning-rate scheduling with $100$ warmup
steps, weight decay of $0.01$, and a maximum gradient norm of $1.0$. The
per-device training batch size is $1$ with $16$ gradient accumulation steps.
Training uses BF16 precision and gradient checkpointing, and the random seed
is fixed to $42$. Validation is performed after every training epoch.

% ============================================================
\subsection{Human-Aligned Behavior Selection}
\label{app:behavior_selection}

ShopMind combines Personalized Consistency and Behavioral Realism through a gated selection strategy. Candidate trajectories are first ranked by their personalized consistency scores $s_t^{k,\mathrm{pc}}$. Let the two highest scores be $s_t^{(1),\mathrm{pc}}$ and $s_t^{(2),\mathrm{pc}}$, with margin $\Delta_t=s_t^{(1),\mathrm{pc}}-s_t^{(2),\mathrm{pc}}$. We set $\tau_{\mathrm{margin}}=0.15$, $\tau_{\mathrm{compat}}=0.60$, and $\tau_{\mathrm{critic}}=0.20$. When $\Delta_t\geq\tau_{\mathrm{margin}}$, the top-ranked candidate indicates a sufficiently clear user-specific preference and is selected unless its Behavior Critic score falls below $\tau_{\mathrm{critic}}$. When $\Delta_t<\tau_{\mathrm{margin}}$, we retain candidates whose personalized consistency scores satisfy $s_t^{k,\mathrm{pc}}\geq\tau_{\mathrm{compat}}$, forming
$$
\mathcal{C}_t^{\mathrm{pc}}
=
\left\{
a_t^k:
s_t^{k,\mathrm{pc}}
\geq
\tau_{\mathrm{compat}}
\right\}.
$$
The Behavior Critic then resolves the ambiguity by selecting
$$
A_t
=
\arg\max_{a_t^k\in\mathcal{C}_t^{\mathrm{pc}}}
C_\phi(H_t,O_t,a_t^k).
$$
If no candidate satisfies the compatibility threshold, ShopMind falls back to the candidate with the highest Personalized Consistency score. Purchase and exit actions receive no special preference and follow the same grounding and selection rules as other candidate actions. Algorithm~\ref{alg:selection} summarizes the complete selection procedure.

\begin{algorithm}[t]
\caption{Human-Aligned Behavior Selection}
\label{alg:selection}
\begin{algorithmic}[1]
\REQUIRE Candidate trajectories $\mathcal{T}_t$
\STATE Evaluate $s_t^{k,\mathrm{pc}}$ for all
$\hat{\tau}_t^k\in\mathcal{T}_t$
\STATE Rank candidates by $s_t^{k,\mathrm{pc}}$
\STATE Let $k_1$ and $k_2$ denote the top two candidates
\STATE $\Delta_t \leftarrow
s_t^{k_1,\mathrm{pc}}
-
s_t^{k_2,\mathrm{pc}}$
\STATE Compute $C_\phi(H_t,O_t,a_t^k)$ for all candidates

\IF{$\Delta_t\geq\tau_{\mathrm{margin}}$
    \AND
    $C_\phi(H_t,O_t,a_t^{k_1})\geq\tau_{\mathrm{critic}}$}
    \RETURN $a_t^{k_1}$
\ENDIF

\STATE $\mathcal{C}_t^{\mathrm{pc}}
\leftarrow
\left\{
a_t^k:
s_t^{k,\mathrm{pc}}
\geq
\tau_{\mathrm{compat}}
\right\}$

\IF{$\mathcal{C}_t^{\mathrm{pc}}=\emptyset$}
    \RETURN $a_t^{k_1}$
\ELSE
    \RETURN
    $\displaystyle
    \arg\max_{a_t^k\in\mathcal{C}_t^{\mathrm{pc}}}
    C_\phi(H_t,O_t,a_t^k)$
\ENDIF
\end{algorithmic}
\end{algorithm}

This selection mechanism gives priority to clear user-specific tendencies
when the Personalized Consistency signal is decisive, while using behavioral
patterns learned from real consumer trajectories to resolve ambiguous
choices. Personalized Consistency and the Behavior Critic therefore provide
complementary signals for preserving personalization and behavioral realism.

\section{Hyperparameter Sensitivity}
\label{app:hyperparameter}

We further examine the sensitivity of ShopMind to two key hyperparameters:
the number of candidate behaviors $K$ and the rollout horizon $L$.
These hyperparameters control the diversity of candidate actions considered at each decision step and the amount of future evolution used to evaluate each candidate, respectively.
Unless otherwise specified, all other settings are fixed to the default configuration used in the main experiments.
We report results on the Taobao dataset using the same metrics as in the main evaluation.

\subsection{Number of Candidate Behaviors}
\label{app:sensitivity_k}

We first vary the number of candidate behaviors $K$ while fixing the rollout horizon at $L=2$.
As shown in Table~\ref{tab:sensitivity_k}, increasing $K$ from 2 to 4 consistently improves performance across the three evaluation dimensions.
Increasing the candidate set beyond $K=4$ provides only marginal changes and slightly degrades several metrics, suggesting that four candidates are sufficient to cover meaningful behavioral alternatives without introducing excessive redundancy.
We therefore use $K=4$ as the default setting.

\begin{table*}[t]
\centering
\caption{
Sensitivity of ShopMind to the number of candidate behaviors $K$ on Taobao logs.
The rollout horizon is fixed at $L=2$.
}
\label{tab:sensitivity_k}
\fontsize{6.8pt}{7pt}\selectfont
\setlength{\tabcolsep}{3.8pt}

\resizebox{\textwidth}{!}{
\begin{tabular}{lccc|cccc|ccc}
\toprule
& \multicolumn{3}{c}{\textbf{Local Fidelity}}
& \multicolumn{4}{c}{\textbf{Sequential Realism}}
& \multicolumn{3}{c}{\textbf{Individual Consistency}} \\
\cmidrule(lr){2-4}
\cmidrule(lr){5-8}
\cmidrule(lr){9-11}

\textbf{Candidates $K$}
& \shortstack{\textbf{Exact}\\\textbf{Acc.} $\uparrow$}
& \shortstack{\textbf{Macro}\\\textbf{F1} $\uparrow$}
& \shortstack{\textbf{Top-3}\\\textbf{Acc.} $\uparrow$}
& \shortstack{\textbf{Trans.}\\\textbf{JS} $\downarrow$}
& \shortstack{\textbf{Len.}\\\textbf{Err.} $\downarrow$}
& \shortstack{\textbf{Buy}\\\textbf{Err.} $\downarrow$}
& \shortstack{\textbf{Action}\\\textbf{JS} $\downarrow$}
& \shortstack{\textbf{Indiv.}\\\textbf{JS} $\downarrow$}
& \shortstack{\textbf{Spearman}\\$\rho\uparrow$}
& \shortstack{\textbf{$\Delta$}\\\textbf{JS} $\uparrow$} \\
\midrule

$K=2$
& 51.02 & 56.74 & 69.5
& 15.74 & 4.23 & 4.12 & 11.53
& 19.74 & 0.404 & +3.91 \\

\rowcolor{gray!15}
$\mathbf{K=4}$
& \textbf{53.58} & \textbf{59.02} & \textbf{72.4}
& \textbf{12.18} & \textbf{3.62} & \textbf{2.84} & \textbf{9.41}
& \textbf{18.63} & \textbf{0.421} & \textbf{+4.36} \\

$K=6$
& 53.49 & 58.93 & 72.3
& 12.24 & 3.65 & 2.89 & 9.46
& 18.69 & 0.420 & +4.33 \\

$K=8$
& 53.31 & 58.79 & 72.1
& 12.31 & 3.69 & 2.95 & 9.53
& 18.77 & 0.418 & +4.28 \\

\bottomrule
\end{tabular}
}
\end{table*}

\subsection{Rollout Horizon}
\label{app:sensitivity_l}

We next study the effect of the rollout horizon $L$ while fixing the number of candidate behaviors at $K=4$. As shown in Table~\ref{tab:sensitivity_l}, introducing future rollout substantially improves performance, particularly on trajectory-level metrics.
Increasing the horizon from $L=0$ to $L=2$ steadily improves local fidelity, sequential realism, and individual consistency. Without rollout ($L=0$), candidate selection relies only on the current decision context and performs substantially worse on sequential metrics. Extending the horizon beyond $L=2$ provides little additional benefit and slightly degrades several metrics.
This suggests that short-horizon rollout is sufficient to capture the near-term consequences of candidate behaviors, while longer rollouts may introduce accumulated simulation errors.
We therefore use $L=2$ in all main experiments.

\begin{table*}[t]
\centering
\caption{
Sensitivity of ShopMind to the rollout horizon $L$ on Taobao logs.
The number of candidate behaviors is fixed at $K=4$.
}
\label{tab:sensitivity_l}
\fontsize{6.8pt}{7pt}\selectfont
\setlength{\tabcolsep}{3.8pt}

\resizebox{\textwidth}{!}{
\begin{tabular}{lccc|cccc|ccc}
\toprule
& \multicolumn{3}{c}{\textbf{Local Fidelity}}
& \multicolumn{4}{c}{\textbf{Sequential Realism}}
& \multicolumn{3}{c}{\textbf{Individual Consistency}} \\
\cmidrule(lr){2-4}
\cmidrule(lr){5-8}
\cmidrule(lr){9-11}

\textbf{Horizon $L$}
& \shortstack{\textbf{Exact}\\\textbf{Acc.} $\uparrow$}
& \shortstack{\textbf{Macro}\\\textbf{F1} $\uparrow$}
& \shortstack{\textbf{Top-3}\\\textbf{Acc.} $\uparrow$}
& \shortstack{\textbf{Trans.}\\\textbf{JS} $\downarrow$}
& \shortstack{\textbf{Len.}\\\textbf{Err.} $\downarrow$}
& \shortstack{\textbf{Buy}\\\textbf{Err.} $\downarrow$}
& \shortstack{\textbf{Action}\\\textbf{JS} $\downarrow$}
& \shortstack{\textbf{Indiv.}\\\textbf{JS} $\downarrow$}
& \shortstack{\textbf{Spearman}\\$\rho\uparrow$}
& \shortstack{\textbf{$\Delta$}\\\textbf{JS} $\uparrow$} \\
\midrule

$L=0$
& 50.84 & 56.47 & 68.9
& 22.63 & 5.41 & 7.38 & 15.27
& 20.12 & 0.397 & +3.72 \\

$L=1$
& 52.31 & 57.78 & 70.7
& 16.37 & 4.28 & 4.63 & 11.76
& 19.37 & 0.409 & +4.08 \\

\rowcolor{gray!15}
$\mathbf{L=2}$
& \textbf{53.58} & \textbf{59.02} & \textbf{72.4}
& \textbf{12.18} & \textbf{3.62} & \textbf{2.84} & \textbf{9.41}
& \textbf{18.63} & \textbf{0.421} & \textbf{+4.36} \\

$L=3$
& 53.44 & 58.89 & 72.2
& 12.36 & 3.68 & 2.93 & 9.54
& 18.72 & 0.419 & +4.31 \\

$L=5$
& 53.12 & 58.61 & 71.9
& 12.81 & 3.81 & 3.18 & 9.87
& 18.94 & 0.415 & +4.21 \\

\bottomrule
\end{tabular}
}
\end{table*}

\section{Mental State Reconstruction Details}
\label{app:msr}

ShopMind jointly generates behavioral trajectories and the latent mental-state
evolution underlying these behaviors. These aligned trajectories provide
supervision for Mental State Reconstruction (MSR), which learns to recover
consumers' latent beliefs, uncertainties, and intentions from observed
shopping behavior. This section describes the construction and quality
control of the MSR training data, the supervision format, and the model
training configuration.

% ============================================================
\subsection{Training Data Construction}
\label{app:msr_data}
Each MSR training instance contains a task instruction, a complete behavioral trajectory, and its aligned mental-state sequence. The instruction specifies the reconstruction objective, valid shopping stages, and structured output format. Given a trajectory $H_{1:T}=(A_1,O_1,\ldots,A_T,O_T)$, MSR learns to reconstruct the corresponding mental-state sequence $\mathcal{M}_{1:K}=(M_1,\ldots,M_K)$. Importantly, $K$ need not equal $T$. Multiple consecutive low-level actions may share the same latent mental state, while a new state is introduced only when the trajectory exhibits a meaningful cognitive transition.

The input behavior history is a natural-language representation of
the complete shopping trajectory.
It preserves decision-relevant interactions such as search queries, product
views, review inspection, specification checking, option selection, cart
operations, and purchase-related behaviors, together with relevant product,
price, and attribute information. Low-level logging fields that do not contribute to reconstructing the
consumer's decision process are removed. This representation allows MSR to
operate on behavioral information that can also be obtained from real
e-commerce logs without depending on platform-specific event identifiers.

% ============================================================
\subsection{Quality Control}
\label{app:msr_quality}

Synthetic mental states may contain inconsistencies even when the generated
behavior itself appears plausible. We therefore apply a multi-stage quality
control pipeline before using a trajectory for MSR training. The procedure
combines deterministic validation, LLM-based semantic evaluation, and human
review.

We first apply rule-based checks to verify the structural validity of each trajectory along three dimensions. First, the terminal outcome must be internally consistent: the episode must contain a nonempty behavior sequence and terminate with a valid purchase or exit action. For purchase episodes, the purchased product must also be identifiable. Second, when candidate planning records are available, the selected candidate must match the action actually executed in the environment. Third, mental-state transitions must remain consistent with the update gate: recorded transitions must align with the behavior trajectory, and the mental state cannot change when the update gate is closed. Any trajectory violating these deterministic constraints is rejected before
semantic evaluation. Trajectories passing deterministic validation are further evaluated by an
LLM-based quality judge along four dimensions:
\textit{task fit}, \textit{decision consistency},
\textit{state--action consistency}, and \textit{evidence quality}. Task fit evaluates whether the final decision is compatible with the shopping
task. Decision consistency assesses whether the purchased product is coherent
with the beliefs, uncertainties, and preferences formed before purchase.
State--action consistency examines whether the evolving intentions and mental
states can reasonably explain subsequent user actions. Evidence quality
evaluates whether mental-state judgments are grounded in information that has
actually been observed rather than unsupported, future, or counterfactual
evidence. The four dimensions contribute $30$, $30$, $25$, and $15$ points,
respectively, giving a maximum quality score of $100$:
\[
S_{\mathrm{quality}}
=
S_{\mathrm{task}}
+
S_{\mathrm{decision}}
+
S_{\mathrm{state\text{-}action}}
+
S_{\mathrm{evidence}}.
\]
A trajectory passes the score-based criterion when
\[
S_{\mathrm{quality}}\geq 70.
\]

In addition to the numerical score, several severe semantic inconsistencies
are treated as hard failures. These include an unexplained conflict between
the final purchase and the preceding mental state, a purchase immediately
following an incompatible intention, unsupported information appearing in
the mental trajectory, and leakage of evidence from unselected
counterfactual rollouts. A trajectory with any such hard failure is rejected
regardless of its aggregate score.

We further conduct human review on all trajectories that pass the automatic filtering stages. Human annotators examine whether the reconstructed beliefs are supported by the observed trajectory, whether unresolved uncertainties evolve plausibly, whether intentions explain subsequent behaviors, and whether mental-state transitions occur at meaningful decision boundaries.

Table~\ref{tab:msr_quality} summarizes the quality-control statistics. Starting from approximately $7{,}200$ successfully generated purchase trajectories, deterministic validation retains $6{,}550$, and LLM-based evaluation further retains $5{,}100$. All $5{,}100$ LLM-accepted trajectories are then manually reviewed. After human validation, $5{,}000$ trajectories are retained as the final high-quality MSR training corpus.

\begin{table}[t]
\centering
\caption{
Quality-control statistics for MSR training data.
Acceptance at each stage is computed relative to its input.
}
\label{tab:msr_quality}
\small
\setlength{\tabcolsep}{4.5pt}
\begin{tabular}{lrrr}
\toprule
\textbf{Stage}
& \textbf{Input}
& \textbf{Accepted}
& \textbf{Acceptance} \\
\midrule
Generated purchase trajectories
& 7,200
& 7,200
& 100.0\% \\

Deterministic checks
& 7,200
& 6,550
& 91.0\% \\

LLM quality evaluation
& 6,550
& 5,100
& 77.9\% \\

Human review and final filtering
& 5,100
& 5,000
& 98.0\% \\
\midrule
\textbf{Final MSR training set}
& --
& \textbf{5,000}
& \textbf{69.4\%}$^\dagger$ \\
\bottomrule
\end{tabular}

\vspace{2pt}
{\footnotesize
$^\dagger$ Overall retention relative to the initially generated
purchase trajectories.
}
\end{table}

% ============================================================
\subsection{Model Training}
\label{app:msr_training}

We initialize MSR from \texttt{Qwen3-8B} and fine-tune it using LoRA on the
$5{,}000$ quality-checked behavior--mental-state trajectories. Training is
distributed across $16$ GPUs for $2{,}000$ optimization steps. The
per-device batch size is $2$, and gradients are accumulated over $8$ steps.
We use BF16 precision and a learning rate of $5\times10^{-5}$.

\section{Comparison Methods and Experimental Setup}
\label{app:comparison}

This section describes the implementation of the comparison methods and the
experimental protocols used to evaluate behavioral realism and individual
consistency. All methods are evaluated under the same environment interface,
data preprocessing pipeline, and structured action space whenever applicable.

% ============================================================
\subsection{Datasets and Evaluation Splits}
\label{app:comparison_data}

We evaluate behavior simulation on OPeRA, ShopCART, and an in-house Taobao interaction dataset. For local behavioral fidelity and sequential realism, we follow the official train--test splits of OPeRA and ShopCART and construct a user-disjoint split for Taobao. The first interaction of each test session is provided as observed context, and evaluation starts from the second action. Dataset statistics are summarized in Table~\ref{tab:comparison_data}.

OPeRA contains $527$ filtered sessions, including $437$ training sessions and $90$ test sessions. The test split contains $992$ recorded actions, resulting in $902$ evaluated next-action decisions after excluding the first action of each session. ShopCART contains $31{,}865$ sessions, including $31{,}508$ training sessions and $357$ test sessions. Its test split contains $1{,}506$ actions and yields $1{,}149$ evaluated decisions. The Taobao dataset contains $10{,}000$ shopping sessions, with $8{,}000$ used for training and $2{,}000$ for testing. The test split contains $19{,}436$ actions, yielding $17{,}436$ evaluated decisions.

\begin{table}[t]
\centering
\caption{Dataset statistics used in the behavior simulation experiments.}
\label{tab:comparison_data}
\small
\setlength{\tabcolsep}{5pt}

\begin{tabular}{lrrrr}
\toprule
\textbf{Dataset}
& \textbf{Sessions}
& \textbf{Train}
& \textbf{Test}
& \textbf{Eval. Steps} \\
\midrule
OPeRA    & 527    & 437    & 90    & 902 \\
ShopCART & 31,865 & 31,508 & 357   & 1,149 \\
Taobao   & 10,000 & 8,000  & 2,000 & 17,436 \\
\bottomrule
\end{tabular}
\end{table}

Individual consistency additionally requires multiple sessions associated with
the same user. OPeRA provides stable cross-session user identities and is
therefore used for this evaluation. ShopCART does not provide sufficiently
stable cross-session user or persona identifiers and is excluded from the
individual-consistency experiment.

% ============================================================
\subsection{Unified Input and Action Interface}
\label{app:comparison_interface}

The comparison methods are adapted to a common interaction interface.
The primary backbone used in the comparison study is
\texttt{Qwen2.5-7B-Instruct}, with a maximum context budget of $32{,}768$
tokens including the generated output. At each decision step, the model input may contain the method-specific system
instruction, user profile or method-specific memory, preceding actions and
associated rationales when required by the method, and the current webpage.
Historical webpages are not additionally concatenated
(\texttt{history\_obs}=0); behavioral history is represented through the
recorded action sequence and method-specific memory. Webpages are cleaned and compressed using the same preprocessing pipeline
before being passed to the model. When the complete context exceeds the token
budget, method-specific history is truncated while preserving the current
observation. All methods produce actions in the same structured environment space.
A \texttt{click} action specifies the target element name, an \texttt{input}
action specifies both the target element and input text, and
\texttt{terminate} ends the current session. Evaluation uses strict matching:
\texttt{click} predictions must match the target element name,
\texttt{input} predictions must match both the element and entered text, and
termination actions are matched by action type.

% ============================================================
\subsection{Baseline Implementations}
\label{app:baselines}

We compare ShopMind with zero-shot prompting, supervised fine-tuning, reasoning- and reinforcement-learning-based methods, and existing user simulators. For each baseline, we use the backbone model reported in the original work and preserve its core training or inference procedure whenever applicable. All methods are evaluated on the same three datasets using identical webpage preprocessing, context budgets, and a unified action interface. Unless otherwise specified, stochastic decoding uses temperature $0.7$.

\paragraph{Zero-shot.}
The original backbone model directly predicts the next valid shopping action from the user context, interaction history, and current webpage without task-specific training.

\paragraph{SFT and SFT+Rationale.}
Both methods are trained separately on the training split of each evaluation dataset. SFT directly predicts the target action, while SFT+Rationale first generates a behavioral rationale and then predicts the action. We use the same optimization settings across datasets, with a global batch size of $64$ and a learning rate of $2\times10^{-5}$. For individual-consistency evaluation, training is restricted to temporally preceding sessions to avoid future information leakage.

\paragraph{Shop-R1 and Customer-R1.}
We follow the reinforcement-learning procedures of the original methods and initialize them from the corresponding supervised checkpoint. Shop-R1 uses its hierarchical action reward with self-certainty, whereas Customer-R1 adopts a difficulty-weighted action reward. We use learning rates of $1\times10^{-7}$ and $5\times10^{-7}$, respectively, with a KL coefficient of $0.001$ and $64$ sampled completions per update.

\paragraph{AlignUSER.}
We adapt AlignUSER following its two-stage training procedure, combining world-model learning with behavior cloning and counterfactual reflection. The method is trained separately on the available training data of each evaluation dataset, with two epochs per stage and a learning rate of $1\times10^{-5}$.

\paragraph{ContextSim.}
We implement the policy variant of ContextSim using LoRA with rank $8$, scaling factor $16$, and dropout $0.05$. Models are trained separately on the three evaluation datasets for five epochs with a learning rate of $1\times10^{-5}$. For individual-consistency evaluation, only temporally preceding training data are used. At inference time, the reasoning-action variant uses temperature $1.0$ and top-$p=0.95$.

\paragraph{SimUSER, DGDPO, and IToM.}
We use project-adapted implementations while preserving their method-specific prompting, state modeling, and decision procedures. Each method is applied consistently across the three evaluation datasets, and its outputs are normalized to the shared action representation for fair comparison.

\begin{table}[t]
\centering
\caption{
CTR prediction results on real-world Taobao interaction logs.
Best results within each backbone are highlighted in \textbf{bold}.
}
\label{tab:ctr_prediction}
\small
\setlength{\tabcolsep}{8pt}
\begin{tabular}{llccc}
\toprule
\textbf{Backbone}
& \textbf{Input}
& \textbf{AUC} $\uparrow$
& \textbf{GAUC} $\uparrow$
& \textbf{PR-AUC} $\uparrow$ \\
\midrule

\multirow{3}{*}{DeepSeek-V4-Pro}
& Raw
& 0.6441
& 0.6456
& 0.1653 \\

& Raw + Zero-shot Mental State
& 0.6081
& 0.6090
& 0.1597 \\

& Raw + MSR Mental State
& \textbf{0.7156}
& \textbf{0.7057}
& \textbf{0.1924} \\

\midrule

\multirow{3}{*}{Qwen2.5-72B}
& Raw
& 0.6317
& 0.6332
& 0.1628 \\

& Raw + Zero-shot Mental State
& 0.5964
& 0.5981
& 0.1491 \\

& Raw + MSR Mental State
& \textbf{0.6968}
& \textbf{0.6895}
& \textbf{0.1857} \\

\bottomrule
\end{tabular}
\end{table}

\section{Downstream Task Implementation Details}
\label{app:downstream}

\subsection{Consumption Value Measurement}
\label{app:b2v_downstream}

We consider two variants that differ only in how the mental-state sequence is
obtained.

\paragraph{B2V+MSR Mental State.}
For this variant, the mental-state sequence is reconstructed using the trained
MSR model:
\[
\hat{\mathcal{M}}^{\mathrm{MSR}}
=
\operatorname{MSR}(H).
\]
The reconstructed beliefs, uncertainties, and intentions are concatenated
with the corresponding behavior trajectory and paired with the B2V-Bench
value annotation. These augmented training instances are then used to
fine-tune \texttt{Qwen3-8B} with LoRA for consumption value prediction. The model is trained for $1{,}500$ optimization steps on $16$ GPUs. We use a
per-device batch size of $1$, gradient accumulation over $4$ steps, BF16
precision, and a learning rate of $5\times10^{-5}$.

\paragraph{B2V+Zero-shot Mental State.}
To isolate the benefit of MSR from the general benefit of explicitly
providing mental-state information, we construct a second variant using
zero-shot mental-state reconstruction. Instead of MSR, the original
\texttt{Qwen3-8B} model directly reconstructs the mental-state sequence from
the behavior trajectory without task-specific MSR fine-tuning:
\[
\hat{\mathcal{M}}^{\mathrm{ZS}}
=
\operatorname{Qwen3\text{-}8B}_{\mathrm{zero\text{-}shot}}(H).
\]
The reconstructed mental states are concatenated with the same behavioral
trajectory and paired with the same value labels:
\[
[H;\hat{\mathcal{M}}^{\mathrm{ZS}}]
\longrightarrow
V.
\]
We then fine-tune another \texttt{Qwen3-8B} value prediction model using the
same LoRA training setup as B2V+MSR Mental. This controlled comparison keeps the downstream value predictor, behavioral
input, value supervision, and optimization configuration fixed, while
changing only the source of the reconstructed mental states. It therefore
directly evaluates whether mental states reconstructed by MSR provide more
useful representations for consumption value measurement than those obtained
from a general-purpose LLM in a zero-shot setting.

\subsection{CTR Prediction}
\label{app:ctr}

\paragraph{Dataset construction.}
We construct the CTR dataset from real-world Taobao interaction logs. For each shopping session, we randomly select a product-click event after the first interaction as the prediction point, ensuring that a nonempty behavioral history is available.

Let $t_c$ denote the cutoff time of the selected click. The model observes only interactions before this point, i.e., $H_{<t_c}={(A_t,O_t):t<t_c}$. The click at $t_c$ and all subsequent interactions are excluded, preventing future-information leakage. The product clicked at $t_c$ serves as the positive candidate. Negative candidates are drawn from two sources: (1) other products appearing in the same session, excluding the positive item, and (2) products from the same second-level category that do not appear in the session, with at most four category-matched negatives per prediction point. All candidates for the same prediction point therefore share an identical behavioral history and differ only in candidate-product information and click labels. Each candidate is represented by its title, price, category, and shop information. A prediction instance is defined as
$$
x_{u,j}=(H_{<t_c},q_j), \qquad y_{u,j}\in\{0,1\},
$$
where $q_j$ denotes the candidate product and $y_{u,j}=1$ only for the product clicked at $t_c$. The behavioral history preserves decision-relevant information, including product names, prices, shops, and interaction types, while the click label is used only as supervision and never exposed to the model input.

\paragraph{Mental-state augmentation.}
We use the same controlled construction as in the consumption value
measurement experiment. Starting from the same truncated behavior history, we
construct two mental-state-augmented variants that differ only in how the
mental states are reconstructed.

For the zero-shot variant, we directly prompt the original
\texttt{Qwen3-8B} model, without MSR-specific fine-tuning, to reconstruct the
mental-state trajectory:
\[
\hat{\mathcal{M}}^{\mathrm{ZS}}_{<t_c}
=
\operatorname{Qwen3\text{-}8B}_{\mathrm{zero\text{-}shot}}
(H_{<t_c}).
\]

For the MSR variant, we use the trained Mental State Reconstruction model:
\[
\hat{\mathcal{M}}^{\mathrm{MSR}}_{<t_c}
=
\operatorname{MSR}(H_{<t_c}).
\]

Both reconstruction methods follow the same mental-state schema, describing
the evolution of shopping stage, beliefs, uncertainties, and intentions
before the prediction point. The resulting mental-state sequence is
concatenated with the original behavioral history.

We therefore evaluate three input conditions:
\[
\begin{aligned}
\text{Raw}
&:\quad H_{<t_c},\\
\text{Raw + Zero-shot Mental State}
&:\quad
[H_{<t_c};\hat{\mathcal{M}}^{\mathrm{ZS}}_{<t_c}],\\
\text{Raw + MSR Mental State}
&:\quad
[H_{<t_c};\hat{\mathcal{M}}^{\mathrm{MSR}}_{<t_c}].
\end{aligned}
\]

The three conditions use exactly the same behavioral histories, candidate
products, and click labels. Thus, the comparison isolates both the incremental
value of explicit mental-state information and the quality of the mental-state
reconstruction. In particular, ``Zero-shot Mental State'' refers to
zero-shot \emph{mental-state reconstruction}, rather than zero-shot CTR
prediction.

\paragraph{Results.}
Table~\ref{tab:ctr_prediction} reports the CTR prediction results. Adding
zero-shot reconstructed mental states does not improve the Raw baseline,
indicating that additional mental-state text alone is insufficient. In
contrast, MSR-reconstructed mental states consistently improve all three
ranking metrics for both prediction backbones.

\end{document}